\documentclass[conference]{IEEEtran}

\usepackage{graphicx}
\usepackage{epsfig}
\usepackage{epstopdf}	    
\usepackage{pslatex}	
\usepackage{url}
\usepackage{amsmath}

\begin{document}

\title{Harnessing the Deep Net Object Models for enhancing Human Action Recognition}
\author{\IEEEauthorblockN{O.V. Ramana Murthy$^1$ and Roland Goecke$^{1,2}$}
\IEEEauthorblockA{$^1$Vision \& Sensing, HCC Lab, ESTeM, University of Canberra\\
$^2$IHCC, RSCS, CECS, Australian National University\\
Email: O.V.RamanaMurthy@ieee.org, roland.goecke@ieee.org}}
\maketitle

\begin{abstract}
In this study, the influence of objects is investigated in the scenario of human action recognition with large number of classes. We hypothesize that the objects the humans are interacting will have good say in determining the action being performed. Especially, if the objects are non-moving, such as objects appearing in the background, features such as spatio-temporal interest points, dense trajectories may fail to detect them. Hence we propose to detect objects using pre-trained object detectors in every frame statically. Trained Deep network models are used as object detectors. Information from different layers in conjunction with different encoding techniques is extensively studied to obtain the richest feature vectors. This technique is observed to yield state-of-the-art performance on \textbf{HMDB51} and \textbf{UCF101} datasets.
\end{abstract}


%
%
\section{Introduction}
\label{sec::Introduction}
We deal with the problem of supervised human action recognition from unconstrained `real-world' videos. The objective is to determine an action (one per time instance) performed in a given video. In the scenario where large number of action classes are present such as \textbf{HMDB51} \cite{HMDB} and \textbf{UCF101} \cite{UCF101} with 51 and 101 classes respectively, five Major categories \cite{HMDB} can be classified as shown in Table \ref{tab::Categories}. The major discriminating information between two categories such as (1) and (2) is the objects with which the human are interacting! The discriminating information for action classes within a category such as in (4) -- \textit{Shoot ball}, \textit{Shoot bow}, \textit{Shoot gun} -- is the objects information. Further, if the objects were to be non-moving, such as \textit{gun}, \textit{bow}, they would not be detected by the spatio-temporal interest points or trajectories. 

To overcome these limitations, we use Convolution Neural Networks (CNN) pre-trained on the Imagenet \cite{ImageNet}
1000 object categories. CNNs are very efficient to train, faster to apply and better in accuracy as objects detectors \cite{ImageNet_CNN}. These deep nets learn the invariant representation
and object classification result simultaneously by back-propagating information, through stacked convolution
and pooling layers, with the aid of a large number of labelled examples.

\begin{table*}[ht]
\caption{Broader Categories in Human Action Recognition}
\begin{center}
\begin{tabular}{ccc}
\hline
Index & Category &  Actions in the \textbf{HMDB51}  dataset \\ 
\hline \hline
1 & General facial actions & Smile, Laugh, Chew, Talk \\
2 & Facial actions with object manipulation & Smoke, Eat, Drink\\
3 & General body movements & Cartwheel, Clap hands, Climb, Climb stairs, Dive, Backhand flip,  Fall on the floor, Handstand,  \\
 & & Jump, Pull up, Push up, Run, Sit down, Sit up, Somersault, Stand up, Turn, Walk, Wave\\
4&Body movements with object interaction & Brush hair, Catch, Draw sword, Dribble, Golf, Hit something,Kick ball, Pick, Pour, Push something,  \\
&& Ride bike, Ride horse, Shoot ball, Shoot bow, Shoot gun, Swing baseball bat, Sword exercise, Throw\\
5&Body movements for human interaction & Fencing, Hug, Kick someone, Kiss, Punch, Shake hands, Sword fight\\
\hline
\end{tabular}
\label{tab::Categories}
\end{center}
\end{table*}

In this context, we investigate the following questions

\begin{enumerate}
\item What is the influence of objects in human action recognition ?
\item Generalization capabilities of the constructed object feature vectors
\end{enumerate}
We study and present our results on the large-scale action datasets \textbf{HMDB51} and \textbf{UCF101} containing at least 51-101 different action classes.

In the remainder of the paper, Section \ref{sec::Literature} contains a review of the related works. Section \ref{sec::framework} describes the framework and details the local feature descriptors, codebook generation, object detectors, different feature encoding techniques, classifier and datasets. Section \ref{sec::Results} presents and discusses the results obtained on the benchmark datasets. Finally, conclusions are drawn in Section \ref{sec::Conclusions}.

%
%
\section{Related Literature}
\label{sec::Literature}
The influence of objects in human action recognition has begun with recent works \cite{Mihir2015} and \cite{ICMR2015}. Jain \textit{et al.} \cite{Mihir2015} conduct an empirical study on the benefit of encoding 15,000 object categories for action recognition. They show that objects matter for actions, and are often semantically relevant as well. And,when objects are combined with motion, improve the state-of-the-art for both action classification and localization. They train 15,000 object classifiers from the Imagenet \cite{ImageNet}
with a deep convolutional neural network \cite{ImageNet_CNN} and use their pooled responses as a video representation for action classification and localization.  However, they utilized only the final output responses (softmax probability scores) of Deep net models as features. Cai \textit{et al.} \cite{ICMR2015} utilized intermediate-level Deep net model outputs ({\fontfamily{qcr}\selectfont Fc6}) as features along with low level features and 1418 Semantic Concept detectors trained from ConceptsWeb \cite{ConceptsWeb}. ConceptsWeb consists of half a million images downloaded from the web, manually annotated and organized in a hierarchical faceted taxonomy. Further, they fuse these features through extensive experimental evaluations and show improved action classification performance. Xu \textit{et al.} \cite{Xu2015} investigated extensively intermediate-level Deep net model outputs ({\fontfamily{qcr}\selectfont Pool5, Fc6, Fc7}) in conjunction with different feature encoding techniques. However, their Deep net models were trained for action recognition, not object detection. 

%
%
\section{Overall Framework and Background}
\label{sec::framework}
\begin{figure*}[ht]
  \centering
  \centerline{\includegraphics[scale=0.80]{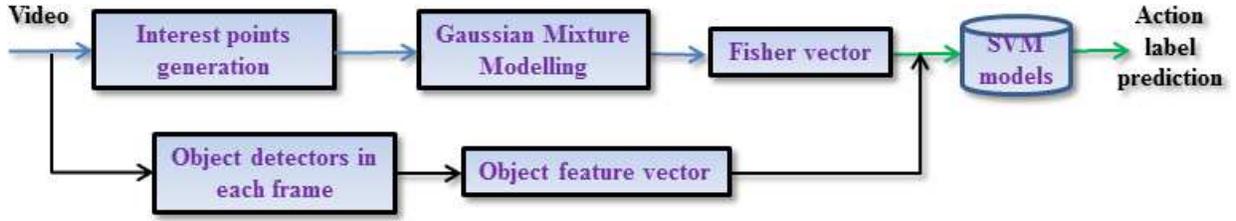}}

\caption{Overall Framework}
\label{fig::OverallLayoutAndFramework}
\end{figure*}

The overall layout of our proposed framework is shown in Fig.\ \ref{fig::OverallLayoutAndFramework}. Firstly, interest points -- trajectories of moving objects -- are detected. Local descriptors are computed around these detected interest points. Gaussian Mixture Modelling is applied and Fisher vectors are generated. In a parallel channel, objects are detected in each frame statically, using pre-trained object detectors from Imagenet. Feature vectors are constructed, using different encoding, from different layers of the pre-trained object detectors. These feature vectors are concatenated to the Fisher vector to learn a classifier (for each action class detection). The details of each stage are discussed in following sections.

%
%

\subsection{Spatio-Temporal Interest Points}
In their seminal work, Laptev \emph{et al.}\ \cite{lapt_lind03} proposed the usage of Harris 3D corners as an extension of traditional (2D) Harris corner points for spatio-temporal analysis and action recognition. These interest points are local maxima of a function of space-time gradients. They compute a spatio-temporal second-moment matrix at each video point in different spatio-temporal scales. This matrix essentially captures space-time gradients. The interest points are obtained as local maxima of a function of this second-moment matrix. We use the original implementation\footnote{\label{STIP}\url{http://www.di.ens.fr/~laptev/download.html/#stip}} with standard parameter settings. These points are extracted at multiple scales based on a regular sampling of spatial and temporal scale values. They are defined in 5 dimensions $(x, y, t, \sigma, \tau)$, where $x$, $y$ and $t$ are spatial and temporal axes, resp., while $\sigma$ and $\tau$ are the spatial and temporal scales, respectively. Local descriptors histograms of oriented gradients (HOG) and histograms of optic flow (HOF) are computed around the detected interest points.

%
%

\subsection{Dense Trajectories}
\label{sec::Trajectories}
Wang \emph{et al.}\ \cite{Trajectories} proposed dense trajectories to model human actions. Interest points were sampled at uniform intervals in space and time, and tracked based on displacement information from a dense optical flow field. Improved dense trajectories (iDT) \cite{Wang2013} are an improved version of the dense trajectories obtained by estimating the camera motion. Wang and Schmid \cite{Wang2013} use a human body detector to separate motion stemming from humans movements from camera motion. The estimate is also used to cancel out possible camera motion from the optical flow. For trajectories of moving objects, we compute these improved dense trajectories. In our experiments, we only use the online version \cite{Wang2013} of camera motion compensated improved trajectories, without any human body detector. The local descriptors computed on these trajectories are HOG, HOF, motion boundary histograms (MBH) and trajectory shape. 

%
%

\subsection{Multi-Skip Feature Stacking}
Generally, action feature extractors such as STIP, iDT, involve differential operators, which act as high-pass filters and tend to attenuate low frequency action information. This attenuation introduces bias to the resulting features and generates ill-conditioned feature matrices. To overcome this limitation, Lan \textit{et al.} \cite{Multi-Skip} proposed Multi-Skip Feature Stacking (MIFS), which stacks features extracted using a family of differential filters parameterized with multiple time skips ($L$) and encodes shift-invariance into the frequency space. MIFS compensates for information lost from using differential operators by recapturing information at coarse scales. This recaptured information matches actions at different speeds and ranges of motion. 

MIFS on improved dense trajectories are used in the current experiments. On the choice of $L$, Lan \textit{et al.} report that having one or two more scales than the original scale is enough to recover most of lost information due to the differential operations. However, higher scale features become less reliable due to the increasing difficulty in optical flow estimation and tracking. Hence, $L=3$ is finalized. 

%
%

\subsection{Object Detectors}
\begin{figure*}[ht]
  \centering
  \centerline{\includegraphics[scale=0.50]{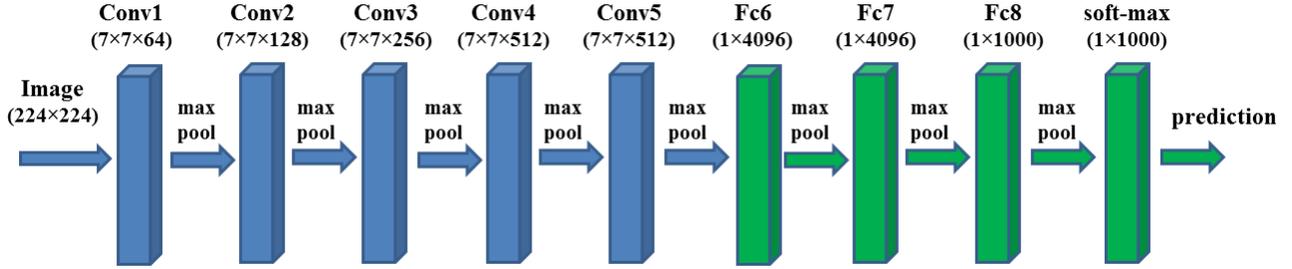}}

\caption{Architecture of {\fontfamily{qcr}\selectfont vgg16}}
\label{fig::vgg16}
\end{figure*}

We utilize existing deep learning framework for computing object detection scores from each video frame. The open source MatConvNet \cite{Matconvnet} implementation based on the deep convolutional neural network architecture by Simonyan and Zisserman \cite{Simonyan14c} is used in all our experiments. We take the ImageNet model, {\fontfamily{qcr}\selectfont Imagenet-vgg-verydeep-16} ,  trained on previous ILSVRC image classification tasks to compute 1000 object detection responses from each frame.  We set the network input to the raw RGB values of the frames, resized to $224 \times 224$ pixels, and the values are forward propagated through 5 convolutional layers (i.e., pooling and ReLU non-linearities) and 3 fully-connected layers (i.e., to determine its final neuron activities). The architecture is shown in Figure \ref{fig::vgg16}.

In Imagenet CNNs, different layers of deep networks can express different information. The fully-connected layer (softmax probabilities) usually denotes high-level concepts. Deeper convolutional layers ({\fontfamily{qcr}\selectfont Fc6, Fc7}) contain global expressions such as object and scene, while shallower convolutional layers contain local characteristics of the image like lines, edges. 
Jain \textit{et al.} \cite{Mihir2015} use fully-connected layers for action recognition; other layers like pooling layer \cite{Xu2015}, convolutional layers \cite{Treasure2015} are also extracted and utilized.  
All these three levels of information are investigated thoroughly in this study and discussed below

\begin{enumerate}

\item \textbf{Objects1K}: This is based on final layer (softmax probabilities). A probability score in the range 0-1 is assigned to each of the 1000 object categories, and totalling to 1. The $N (=1000)$ dimensional vector of object attribute scores $(S(i); i = 1...N)$ is computed for each frame. These vectors are then 
simply averaged across the frame to yield 

\begin{equation}
\textbf{Objects1K} = \dfrac{1}{F} \sum S_{xf}
\end{equation} 

where $F$ is the number of frames in video $x$, $S_{xf}$ is the object vector representation per frame. Jain \textit{et al.} \cite{Mihir2015} used the same strategy -- computing scores for  15K objects. 

\item \textbf{{\fontfamily{qcr}\selectfont Fc6, Fc7} layers}:  The activations of the neurons in the intermediate hidden layers  -- {\fontfamily{qcr}\selectfont Fc6, Fc7} -- can be used as strong features because they contain much richer and more complex representations. Each one of them is of the $4096 D$ dimensions. The response from each frame are average pooled \cite{ICMR2015}, VLAD and Fisher encoded \cite{Xu2015}. Details of VLAD and FV encoding are presented in Section \ref{sec::Feature Encoding}. As the dimensions of descriptors is very high, PCA is applied to reduce to 256D before applying VLAD/Fisher encoding. 

\item  \textbf{Latent Concept Descriptors (LSD)}: Compared to the fully-connected layers, {\fontfamily{qcr}\selectfont pool5} contains
spatial information. The feature dimension of {\fontfamily{qcr}\selectfont pool5} is $a \times a  \times M$, where $a$ is the size of filtered images of the last pooling layer and $M$ is the number of convolutional filters in the last convolutional layer (in our case based on the VGG Imagenet model \cite{Simonyan14c}, $a = 7$ and $M = 512$). Flattening {\fontfamily{qcr}\selectfont pool5} into a vector will yield to very high dimensional features, which will induce heavy computational cost and instability problems \cite{Douze2013}. However, the convolutional filters can be regarded as generalized linear classifiers on the underlying data patches, and each convolutional filter corresponds to a latent concept. Xu \textit{et al.} \cite{Xu2015} formulate the general features from {\fontfamily{qcr}\selectfont pool5} as the vectors of latent concept descriptors, in which each dimension of the latent concept descriptors represents the response of the specific latent concept. Each filter in the last convolutional layer is independent from other filters. The response of the filter is the prediction of the linear classifier on the convolutional location for the corresponding latent concept. In that way, {\fontfamily{qcr}\selectfont pool5} layer of size $a \times a \times M$ is converted into $a^2$ latent concept descriptors with $M$ dimensions. Each latent concept descriptor represents the responses from the $M$ filters for a specific pooling
location. In this case, each frame contains $a^2$ descriptors instead of one descriptor for the frame.  After the latent concept descriptors for all the frames in a video are obtained, PCA is applied to reduce the descriptor dimensions to half (i.e. 256). Then an encoding method -- VLAD and FV -- is applied  to generate the feature vector. 

\end{enumerate}

%
%

\subsection{Feature Encoding}
\label{sec::Feature Encoding}
After the descriptors are obtained, either from low-level features or from CNNs, they have to be encoded to yield the feature vector for each video. Two types of encoding popular in practise in this domain -- Vector of Locally Aggregated Descriptors (VLAD) and Fisher Vector (FV) -- are reviewed and applied in this study.  These techniques are based on a measure determining how much a descriptor belongs to a particular (assigned) visual word. 

\begin{itemize}

\item Vector of Locally Aggregated Descriptors (VLAD)

In this type of encoding, the difference between the descriptors and the closest visual word is collected as residual vectors. K coarse centers ${mu_1,mu_2,...mu_K}$ are generated by K-means clustering from a randomnly selected (100,000) descriptors. For each coarse center dimension $d$ (dimension of the local feature descriptor, $x_i$), a sub-vector $v^i$ is obtained by accumulating the residual vectors as
\begin{equation}
v^i = \sideset{}{}\sum_{x:q(x)=\mu_i}x-\mu
\end{equation}

The obtained sub-vectors are concatenated to yield a $D$-dimensional vector, where $D=k \times d$. Then intra-normalization \cite{VLAD2013} is applied. Further, a two-stage normalisation is applied. Firstly, the `power-law normalisation' \cite{Fisher} is applied. It is a component-wise non-linear operation. Each component $v_j,j=1$ to $D$ is modified as 

\begin{equation}
\label{power}
v_j = {|v_j|}^\alpha \times sign(v_j),
\end{equation}
where $\alpha$ is a parameter such that $\alpha\le 1$. In all experiments, $k = 256$ and $\alpha= 0.2$. Secondly, the vector is $L_2$-normalised as $v= \frac{v}{||v||}$ to yield the VLAD vector.

\item Fisher Vector Encoding

In this technique, a Gaussian Mixture Model (GMM) is fitted to a randomly selected (250,000) descriptors from the training set. Let the parameters obtained from the GMM fitting be defined as $\theta = (\mu_k,\sum_k,\pi_k; k = 1,2,...,K)$ where $\mu_k, \sum_k$ and $\pi_k$ are the mean, covariance and prior probability of each distribution, respectively. The GMM associates each descriptor $X_i$ to a mode $k$ in the mixture with a strength given by the posterior probability
\begin{equation}
q_{ik}=\dfrac{exp[-\dfrac{1}{2}(X_i-\mu_k)^T\sum_{k}^{-1}(X_i-\mu_k)]}{\sum_{t=1}^K exp[-\dfrac{1}{2}(X_i-\mu_t)^T \sum_k^{-1}(X_i-\mu_t)]}
\end{equation}
The mean ($u_{jk}$) and deviation vectors ($v_{jk}$) for each mode \emph{k} are computed as
\begin{equation}
u_{jk} = \dfrac{1}{N\sqrt{\pi_k}} \sum_{i=1}^{N} q_{ik} \frac{x_{ji} - \mu_{jk}}{\sigma_{jk}}
\end{equation}
\begin{equation}
v_{jk} = \dfrac{1}{N \sqrt{2 \pi_k}}\sum_{i=1}^{N} q_{ik} [ (\frac{x_{ji} - \mu_{jk}}{\sigma_{jk}})^2 - 1]
\end{equation}
where $j=1,2,...,D$ spans the local descriptor vector dimensions. The FV is then obtained by concatenating the vectors ($u_{jk}$) and ($v_{jk}$) for each of the \emph{K} modes in the Gaussian mixtures. Similar to VLAD encoding, the FV is also finally normalized by the `power-law normalisation' and $L_2$-normalisation. We concatenate all the Fisher Vectors (of different descriptors) to yield the final feature vector for a given video. 
\end{itemize}

For classification we use the final feature vectors and linear SVM \cite{liblinear}. We apply the one-versus-all approach in all the cases and select the class with the highest score.

%
%

\subsection{Datasets}
\label{Datasets}

We applied our proposed technique on three benchmark datasets: \textbf{HMDB51} \cite{HMDB} and \textbf{UCF101} \cite{UCF101}. \textbf{HMDB51} contains 51 actions categories. Digitised movies, public databases such as the Prelinger archive, videos from YouTube and Google videos were used to create this dataset. For evaluation purposes, three distinct training and testing splits were specified in the dataset. These splits were built to ensure that clips from the same video were not used for both training and testing. For each action category in a split, 70 training and 30 testing clips indices were fixed so that they fulfil the 70/30 balance for each meta tag.  \textbf{UCF101} data set consists of 101 action categories, collected from realistic action videos, e.g.\ from YouTube. Three train-test splits were provided for consistency in reporting performance.

%
%
\begin{table*}[ht]
\caption{Performance of Objects}
\begin{center}
\begin{tabular}{|c|c|c|c|c|}
\hline
Index & Feature type/technique & Dimesnions &\textbf{HMDB51} & \textbf{UCF101} \\
\hline
\hline
i 	& {Objects1K} 					& 1000& 26.0\%& 58.3\% \\
\hline
ii 	& {\fontfamily{qcr}\selectfont Fc6} Average pooling			& 4096&35.1\%  & 70.5\%	\\
 	& {\fontfamily{qcr}\selectfont Fc7} Average pooling			& 4096&32.8\%  & 65.1\%	\\
 	& {\fontfamily{qcr}\selectfont Fc6 + Fc7} Average pooling		& 8192& 35.8\% & 71.7\%\\
\hline
iii	& {\fontfamily{qcr}\selectfont Fc6} Fisher vector				& 130K& 24.4\%  & 58.9\%	\\
 	& {\fontfamily{qcr}\selectfont Fc7} Fisher vector				& 130K&23.9\% & 58.9\%	\\
 	& {\fontfamily{qcr}\selectfont Fc6 + Fc7} Fisher vector		& 130K&24.6\% & 59.1\%\\
\hline
iv 	& {\fontfamily{qcr}\selectfont Fc6} VLAD				&65K& 22.7\% & 53.5\%	\\
 	& {\fontfamily{qcr}\selectfont Fc7} VLAD 				&65K& 28.0\% & 53.8\%	\\
 	& {\fontfamily{qcr}\selectfont Fc6 + Fc7} VLAD 		& 65K&22.0\% & 53.8\%\\
\hline
v 	& LSD Fisher vector			& 130K &  37.8\%  & 71.9\%	\\
 	& LSD VLAD 	 		&65K&\textbf{41.9\%}  & \textbf{78.2\%} \\
\hline
\hline
vi & 15000 objects scores aggregated \cite{Mihir2015}	 & 15K&38.9\% 	& 65.6\% \\
vii& 1000 objects {\fontfamily{qcr}\selectfont Fc6} average pooling \cite{ICMR2015} 	& 4096&33.05\% 	& 65.88\% \\
\hline
\end{tabular}
\label{tab::Objects}
\end{center}
\end{table*}

\section{Results and Discussions}
\label{sec::Results}
We now present the results obtained by applying the techniques discussed earlier. In each Table, the highest recognition rates observed are highlighted.

\subsection{What is the influence of objects in human action recognition ?}

The results obtained by applying CNN models as object detectors are shown in Table \ref{tab::Objects}. Information from different layers of the Deep net models are encoded using different techniques. It is observed that ({\fontfamily{qcr}\selectfont pool5} layer outputs with VLAD encoding technique performed best.

%
%

\begin{table*}[ht]
\caption{Performance of Objects in combination with other feature representations}
\begin{center}
\begin{tabular}{|c|c|c|c|}
\hline
Index & Feature type/technique & \textbf{HMDB51} & \textbf{UCF101} \\
\hline
\hline
i 	& Fisher vector (FV) from STIP \cite{DICTA2014} 		& 37.9\%		& 69.4\% \\
ii 	& STIP + Objects1K 						& 43.1\%		& 75.6\% \\
iii & STIP + ({\fontfamily{qcr}\selectfont Fc6+Fc7}) average pooling		& 42.7\% 		& 80.6\%\\
iv & STIP + LSD Fisher		& 47.5\% 		& 81.1\%\\
v & STIP + LSD VLAD		& 50.5\% 		&84.5\% \\
\hline
vi 	& FV from Improved dense trajectories (iDT)		& 55.9\% \cite{Wang2013}	& 84.8\%\cite{THUMOS} \\
vii & iDT + Objects1K 						& 59.5\% 		& 87.0\%\\
viii & iDT + ({\fontfamily{qcr}\selectfont Fc6+Fc7}) average pooling		& 60.3\%		& 89.4\%\\
ix& iDT + LSD Fisher	&  60.0\% 		& 89.0\%\\
x& iDT + LSD VLAD		& 61.7\% 		& 90.8\%\\
\hline
xi 	& Multi-Skip ($L = 3$) \cite{Multi-Skip}	& 65.1\% 		& 89.1\%\\
xii & Multi-Skip + Objects1K 				& 66.5\%		& 89.3\%\\
xiii& Multi-Skip + ({\fontfamily{qcr}\selectfont Fc6+Fc7}) average pooling&	{ 66.6\%}	& {91.0\%}\\
xiv& Multi-Skip + LSD Fisher		&  66.4\%		& 90.6\% \\
xv& Multi-Skip + LSD VLAD		&  \textbf{68.0\%}		& \textbf{91.9\%} \\
\hline
\end{tabular}
\label{tab::General}
\end{center}
\end{table*}

\subsection{Generalization capabilities of the constructed object feature vectors}
The three best performing features observed in Table \ref{tab::Objects} were found to be  LSD VLAD,  LSD Fisher and ({\fontfamily{qcr}\selectfont Fc6+Fc7}) average pooling. The influence of these features on three different state-of-the-art feature representations are presented in Table \ref{tab::General}. The constructed object feature vectors are complimentary with all the three different kinds of feature representations. This shows that Deep net features are not over-fitted to some databases; yet have generalization capacity. An improvement of 12-14.9\% (absolute), 5.8\% (absolute) and 2.9\%(absolute) has been observed in STIP (Table \ref{tab::General} v), iDT (Table \ref{tab::General} x) and Multi-Skip (iDT) feature representations (Table \ref{tab::General} xv) respectively.

%
%
\begin{table*}[ht]		
\caption{Comparison with other techniques}
\begin{center}
\begin{tabular}{|c|c|c|c|}
\hline
Approach & Brief description&  HMDB51 & UCF101 \\
\hline \hline
Improved dense trajectories (iDT) & Fisher vector (FV)  & 55.9\% \cite{Wang2013}& 84.8\%\cite{THUMOS}\\
\hline
\textbf{Best of Proposed technique}&  Multi-Skip (iDT) + LSD VLAD & \textbf{68.0\%} &  \textbf{91.9\%}\\
\hline
Jain \textit{et al.} 2015 \cite{Mihir2015} & 15K Objects prob scores + iDT & 61.4\% & 88.5\%\\
\hline
Cai \textit{et al.} 2015 \cite{ICMR2015} & Objects {\fontfamily{qcr}\selectfont Fc6} + iDT + semantic concepts & 62.9\% & 89.6\%\\
\hline
Wang \textit{et al.} 2015 \cite{TDD} & Trajectory pooled Descriptors + iDT & 65.9\% & 91.5\%\\
\hline
Miao \textit{et al.} 2015 \cite{TVA2015} & Temporal variance analysis on iDT & 66.4\% & 90.2\%\\
\hline
Lan 2015 \cite{Lan2015}& MIFS(iDT) + ConvISA + MIR & 67.0\% & 90.2\%\\
\hline
\end{tabular}
\label{tab::Fullframe}
\end{center}
\end{table*}

\subsection{Comparison with state-of-the-art}
We compare our results with recent works from 2015 only (as their performances are already better than most of the earlier works). 
The closest works on Deep net based objects influence are by Jain \textit{et al.} \cite{Mihir2015} and Cai \textit{et al.} \cite{ICMR2015}. However, they did not extensively focus on tapping the information from the Deep net models. Jain \textit{et al.} used only final output layer (softmax probability scores) of the Deep net models. {\fontfamily{qcr}\selectfont pool5} layer from Deep net models of 15,000 object detectors might significantly improve the performance. Cai \textit{et al.} used only {\fontfamily{qcr}\selectfont Fc6} information with average pooling. However, {\fontfamily{qcr}\selectfont pool5} layer in conjunction with their Semantic web concepts detectors might also yield improved results. We would like to investigate them in future.

Other latest works, but not focused exclusively on objects are as follows. Wang \textit{et al.} \cite{TDD} learn discriminative convolutional feature maps and conduct trajectory-constrained pooling to aggregate these convolutional features into effective descriptors called as trajectory-pooled deep convolutional descriptor (TDD). Fisher vectors are then constructed from these TDDs.
Miao \textit{et al.} \cite{TVA2015} proposed temporal variance analysis (TVA) as a generalization to better utilize temporal information. TVA learns a linear transformation matrix that projects multidimensional temporal data to temporal components with temporal variance. By mimicking the function of visual cortex (V1) cells, appearance and motion information are obtained by slow and fast features from gray videos using slow and fast filters, respectively. Additional motion features are extracted from optical flows. In this way, slow features
encode velocity information, and fast features encode acceleration information. By using parts of fast filters as slow filters and vice versa, the hybrid slim filter is proposed to improve both slow and fast feature extraction. Finally, they separately encode extracted local features with different temporal variances and concatenate all the encoded features as final features.  Lan \cite{Lan2015} used four complementary methods to improve the performance of action recognition by unsupervised learning from iDT features. Initially, MIFS enhanced iDT features are used to learn motion descriptors using Stacked Convolutional Independent Subspace Analysis (ConvISA) \cite{ConvISA}. Then, spatio-temporal information is incorporated into the learned descriptors by augmenting with normalized spatio-temporal location information. Finally, the relationship among action classes is captured by a Multi-class Iterative Re-ranking (MIR) method \cite{MIR2014} that exploits the relationship among classes. The best Deep net based object information investigated in this paper may also be turn out to be complimentary with these latest methods. This will be investigated in the future work.
 
%
%

\section{Conclusions}
\label{sec::Conclusions}
In this study, the influence of objects using Deep net models was investigated thoroughly. Information from different layers of the the Deep net models was extensively investigated in conjunction with different feature encoding techniques. Information from {\fontfamily{qcr}\selectfont pool5} with VLAD encoding technique was found to be very rich and complementary to different types of low-level feature representations; supporting its generalization capacity. Competitive state-of-the-art performances are achieved on two benchmark datasets. 

%
%

\bibliographystyle{hieeetr}
\bibliography{Rama}

\begin{thebibliography}{10}

\bibitem{HMDB}
H.~Kuehne, H.~Jhuang, E.~Garrote, T.~Poggio, and T.~Serre, ``{HMDB: A large
  video database for human motion recognition},'' in {\em International
  Conference on Computer Vision (ICCV)}, 2011.

\bibitem{UCF101}
K.~Soomro, A.~R. Zamir, and M.~Shah, ``{UCF101: A Dataset of 101 Human Action
  Classes from Videos in the Wild},'' in {\em CRCV-TR-12-01}, November 2012.

\bibitem{ImageNet}
J.~Deng, W.~Dong, R.~Socher, L.-J. Li, K.~Li, and L.~Fei-Fei, ``Imagenet: A
  large-scale hierarchical image database,'' in {\em IEEE Conference on
  Computer Vision and Pattern Recognition (CVPR)}, pp.~248--255, June 2009.

\bibitem{ImageNet_CNN}
A.~Krizhevsky, I.~Sutskever, and G.~E. Hinton, ``Imagenet classification with
  deep convolutional neural networks,'' in {\em Advances in Neural Information
  Processing Systems (NIPS) 25} (F.~Pereira, C.~Burges, L.~Bottou, and
  K.~Weinberger, eds.), pp.~1097--1105, 2012.

\bibitem{Mihir2015}
M.~Jain, J.~van Gemert, and C.~Snoek, ``What do 15,000 object categories tell
  us about classifying and localizing actions?,'' in {\em IEEE Conference on
  Computer Vision and Pattern Recognition (CVPR)}, pp.~46--55, June 2015.

\bibitem{ICMR2015}
J.~Cai, M.~Merler, S.~Pankanti, and Q.~Tian, ``Heterogeneous semantic level
  features fusion for action recognition,'' in {\em Proceedings of the 5th ACM
  International Conference on Multimedia Retrieval (ICMR)}, pp.~307--314, 2015.

\bibitem{ConceptsWeb}
M.~Merler, B.~Huang, L.~Xie, G.~Hua, and A.~Natsev, ``Semantic model vectors
  for complex video event recognition,'' {\em IEEE Transactions on Multimedia},
  vol.~14, pp.~88--101, Feb 2012.

\bibitem{Xu2015}
Z.~Xu, Y.~Yang, and A.~G. Hauptmann, ``A discriminative cnn video
  representation for event detection,'' in {\em IEEE Conference on Computer
  Vision and Pattern Recognition (CVPR)}, pp.~1798--1807, 2015.

\bibitem{lapt_lind03}
I.~Laptev and T.~Lindeberg, ``{Space-Time Interest Points},'' in {\em
  International Conference on Computer Vision (ICCV)}, pp.~432--439, Oct 2003.

\bibitem{Trajectories}
H.~Wang, A.~Kl\"{a}ser, C.~Schmid, and C.-L. Liu, ``{Action recognition by
  dense trajectories},'' in {\em IEEE Conference on Computer Vision and Pattern
  Recognition (CVPR)}, 2011.

\bibitem{Wang2013}
H.~Wang and C.~Schmid, ``{Action Recognition with Improved Trajectories},'' in
  {\em Intenational Conference on Computer Vision (ICCV)}, 2013.

\bibitem{Multi-Skip}
Z.-Z. Lan, M.~Lin, X.~Li, A.~G. Hauptmann, and B.~Raj, ``Beyond gaussian
  pyramid: Multi-skip feature stacking for action recognition.,'' in {\em IEEE
  Conference on Computer Vision and Pattern Recognition (CVPR)}, pp.~204--212,
  2015.

\bibitem{Matconvnet}
A.~Vedaldi and K.~Lenc, ``Matconvnet -- convolutional neural networks for
  matlab,''

\bibitem{Simonyan14c}
K.~Simonyan and A.~Zisserman, ``Very deep convolutional networks for
  large-scale image recognition,'' {\em CoRR}, vol.~abs/1409.1556, 2014.

\bibitem{Treasure2015}
L.~Liu, C.~Shen, and A.~{van den Hengel}, ``The treasure beneath convolutional
  layers: cross convolutional layer pooling for image classification,'' in {\em
  IEEE Conference on Computer Vision and Pattern Recognition (CVPR)}, 2015.

\bibitem{Douze2013}
M.~Douze, J.~Revaud, C.~Schmid, and H.~Jegou, ``Stable hyper-pooling and query
  expansion for event detection,'' in {\em IEEE International Conference on
  Computer Vision (ICCV)}, pp.~1825--1832, Dec 2013.

\bibitem{VLAD2013}
R.~Arandjelovi\'c and A.~Zisserman, ``All about {VLAD},'' in {\em IEEE
  Conference on Computer Vision and Pattern Recognition (CVPR)}, 2013.

\bibitem{Fisher}
F.~Perronnin, J.~S\'{a}nchez, and T.~Mensink, ``{Improving the Fisher kernel
  for large-scale image classification},'' in {\em European Conference on
  Computer Vision (ECCV)}, 2010.

\bibitem{liblinear}
R.-E. Fan, K.-W. Chang, C.-J. Hsieh, X.-R. Wang, and C.-J. Lin, ``{LIBLINEAR}:
  A library for large linear classification,'' {\em Journal of Machine Learning
  Research}, vol.~9, pp.~1871--1874, 2008.

\bibitem{DICTA2014}
O.~V.~R. Murthy and R.~Goecke, ``{The Influence of Temporal Information on
  Human Action Recognition with Large Number of Classes},'' in {\em
  International Conference on Digital Image Computing: Techniques and
  Applications (DICTA)}, 2014.

\bibitem{THUMOS}
H.~Wang and C.~Schmid, ``{LEAR-INRIA submission for the THUMOS workshop},'' in
  {\em THUMOS:ICCV Challenge on Action Recognition with a Large Number of
  Classes, Winner}, 2013.

\bibitem{TDD}
L.~Wang, Y.~Qiao, and X.~Tang, ``Action recognition with trajectory-pooled
  deep-convolutional descriptors.,'' in {\em IEEE Conference on Computer Vision
  and Pattern Recognition (CVPR)}, pp.~4305--4314, 2015.

\bibitem{TVA2015}
J.~Miao, X.~Xu, S.~Qiu, C.~Qing, and D.~Tao, ``Temporal variance analysis for
  action recognition,'' {\em IEEE Transactions on Image Processing}, vol.~24,
  pp.~5904--5915, Dec 2015.

\bibitem{Lan2015}
Z.~Lan, ``Learn to recognize actions through neural networks,'' in {\em
  Proceedings of the 23rd ACM International Conference on Multimedia},
  pp.~657--660, 2015.

\bibitem{ConvISA}
Q.~Le, W.~Zou, S.~Yeung, and A.~Ng, ``Learning hierarchical invariant
  spatio-temporal features for action recognition with independent subspace
  analysis,'' in {\em IEEE Conference on Computer Vision and Pattern
  Recognition (CVPR),}, pp.~3361--3368, June 2011.

\bibitem{MIR2014}
M.~Hoai and A.~Zisserman, ``Improving human action recognition using score
  distribution and ranking,'' in {\em Asian Conference on Computer Vision
  (ACCV)}, pp.~3--20, 2014.

\end{thebibliography}
\end{document}